\documentclass{article}

     \usepackage[preprint]{neurips_2019}

\usepackage[utf8]{inputenc} 
\usepackage[T1]{fontenc}    
\usepackage{hyperref}       
\usepackage{url}            
\usepackage{booktabs}       
\usepackage{amsfonts}       
\usepackage{nicefrac}       
\usepackage{microtype}      

\usepackage{amsmath}
\usepackage{amsfonts}
\usepackage{amssymb} 
\usepackage{array}
\usepackage{caption,subcaption}
\usepackage[enable]{easy-todo}
\usepackage[boxed,vlined,linesnumbered,resetcount,algosection]{algorithm2e}
\usepackage{float}
\usepackage{framed}

\usepackage{tikz}
\usetikzlibrary{patterns,positioning,shadows,arrows,decorations.pathreplacing,shapes.geometric,matrix}

\newcolumntype{x}[1]{%
>{\centering\hspace{0pt}}p{#1}}%

\title{Siamese recurrent networks learn first-order logic reasoning and exhibit zero-shot compositional generalization}

\author{Mathijs Mul \\
  University of Amsterdam \\
  \texttt{mathijsmul@gmail.com} \\\And
  Willem Zuidema \\
  ILLC, University of Amsterdam \\
  \texttt{w.h.zuidema@uva.nl} \\}

\begin{document}

\maketitle

\begin{abstract}
Can neural nets learn logic? We approach this classic question with current methods, and demonstrate that recurrent neural networks can learn to recognize first order logical entailment relations between expressions. We define an artificial language in first-order predicate logic, generate a large dataset of sample `sentences', and use an automatic theorem prover to infer the relation between random pairs of such sentences. We describe a Siamese neural architecture trained to predict the logical relation, and experiment with recurrent and recursive networks. Siamese Recurrent Networks are surprisingly successful at the entailment recognition task, reaching near perfect performance on novel sentences (consisting of known words), and even outperforming recursive networks. We report a series of experiments to test the ability of the models to perform compositional generalization. In particular, we study how they deal with sentences of unseen length, and sentences containing unseen words. We show that set-ups using LSTMs and GRUs obtain high scores on these tests, demonstrating a form of compositionality.

\end{abstract}

\section{Introduction \& related work}

State-of-the-art models for almost all popular natural language processing tasks are based on deep neural networks, trained on massive amounts of data.
A key question that has been raised in many different forms is to what extent these models have learned the compositional generalizations that characterize language, and to what extent they rely on storing massive amounts of exemplars and only make `local' generalizations 
\citep{pinker1988language, fodor1988connectionism, marcus1998rethinking, lake2017building, lake2017still, zhang2016understanding, krueger2017deep, marcus2018deep}.
%
%
This question has led to (sometimes heated) debates between deep learning enthusiasts that are convinced neural networks can do almost anything, and skeptics that are convinced some types of generalization are fundamentally beyond reach for deep learning systems, pointing out that crucial tests distinguishing between generalization and memorization have not been applied. 

In this paper, we take a pragmatic perspective on these issues. As the target for learning we use entailment relations in an artificial language, defined using first order logic (FOL), that is unambiguously compositional. We ask whether popular deep learning methods are capable \emph{in principle} of acquiring the compositional rules that characterize it, and focus in particular on recurrent neural networks that are unambiguously `connectionist': trained recurrent nets do not rely on symbolic data and control structures such as trees and global variable binding, and can straightforwardly be implemented in biological networks \citep{eliasmith2013build} or neuromorphic hardware \citep{Merolla668}.
We report positive results on this challenge, and in the process develop a series of tests for compositional generalization that address the concerns of deep learning skeptics. 

The paper makes three main contributions. First, we develop a protocol for automatically generating data that can be used in entailment recognition tasks. Second, we demonstrate that several deep learning architectures succeed at one such task. Third, we present and apply a number of experiments to test whether models are capable of compositional generalization.


\paragraph{Related work}

Data-driven models have proven successful in various entailment recognition tasks \citep{baroni2012entailment, socher2012, rocktaschel2014low, bowman2014recursive, rocktaschel2015reasoning}.
The data sets used in research on this topic tend to be either fully formal, focusing on logic instead of natural language \citep{evans2018can, allamanis2016learning}, or fully natural, as is the case for manually annotated data sets of English sentence pairs such as SICK \citep{marelli2014semeval} or SNLI \citep{bowman2015large}.
Moreover, entailment recognition models are often endowed with functionality reflecting pre-established linguistic or semantic regularities of the data \citep{bankova2016graded, serafini2016logic, sadrzadeh2018sentence}.
Recently, \citet{shen2018ordered} showed that recurrent networks can learn to recognize logical inference relations if they are extended with a bias towards modelling hierarchical structures.

In this research we do not approach entailment as something fully natural or fully formal, but as a semantic phenomenon that can be recognized in language but that is produced by logic. This perspective was also taken by \citet{bowman2014recursive}, who used a natural logic calculus to infer the entailment relations between pairs of sentences in an artificial language. As opposed to \citeauthor{bowman2014recursive}, we do not use natural logic, which is incomplete and not provably sound, but classical first-order logic (FOL). Furthermore, \citeauthor{bowman2014recursive} used recursive neural networks, shaped according to the syntactic structure of the input sentences, whereas we focus on recurrent networks that receive no linguistic information, and that have no explicit bias to accommodate syntactic hierarchies.

\section{Task definition \& data generation}

The data generation process is inspired by \citet{bowman2014recursive}: an artificial language is defined, sentences are generated according to its grammar and the entailment relation between pairs of such sentences is established according to a fixed background logic. However, our language is significantly more complex, 
and instead of natural logic we use FOL. 


\paragraph{Language}

\begin{figure}[h]
\centering
\begin{subfigure}{0.49\textwidth}
\includegraphics[width=0.95\textwidth]{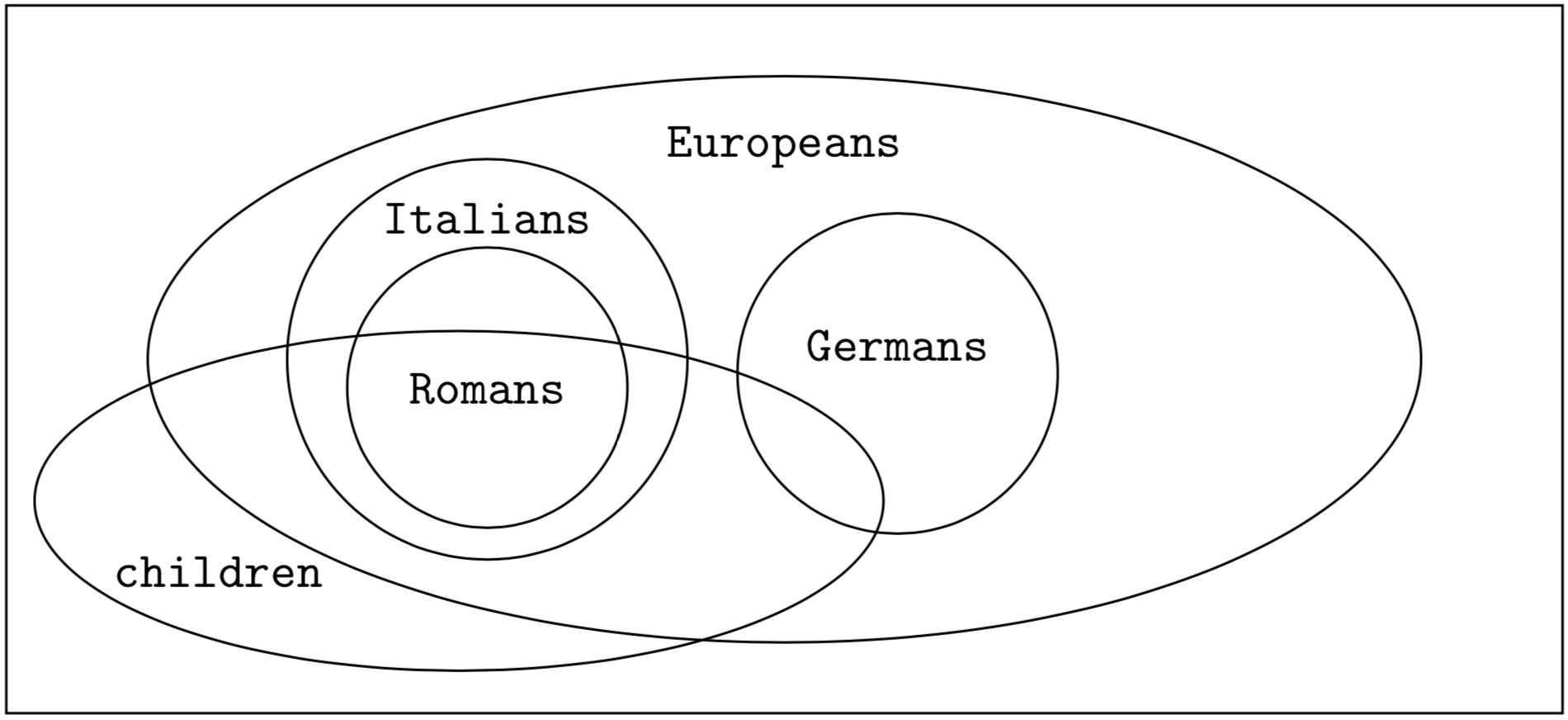}
\caption{}
\label{fig:nouns_new}
\end{subfigure}
\begin{subfigure}{0.49\textwidth}
\includegraphics[width=0.95\textwidth]{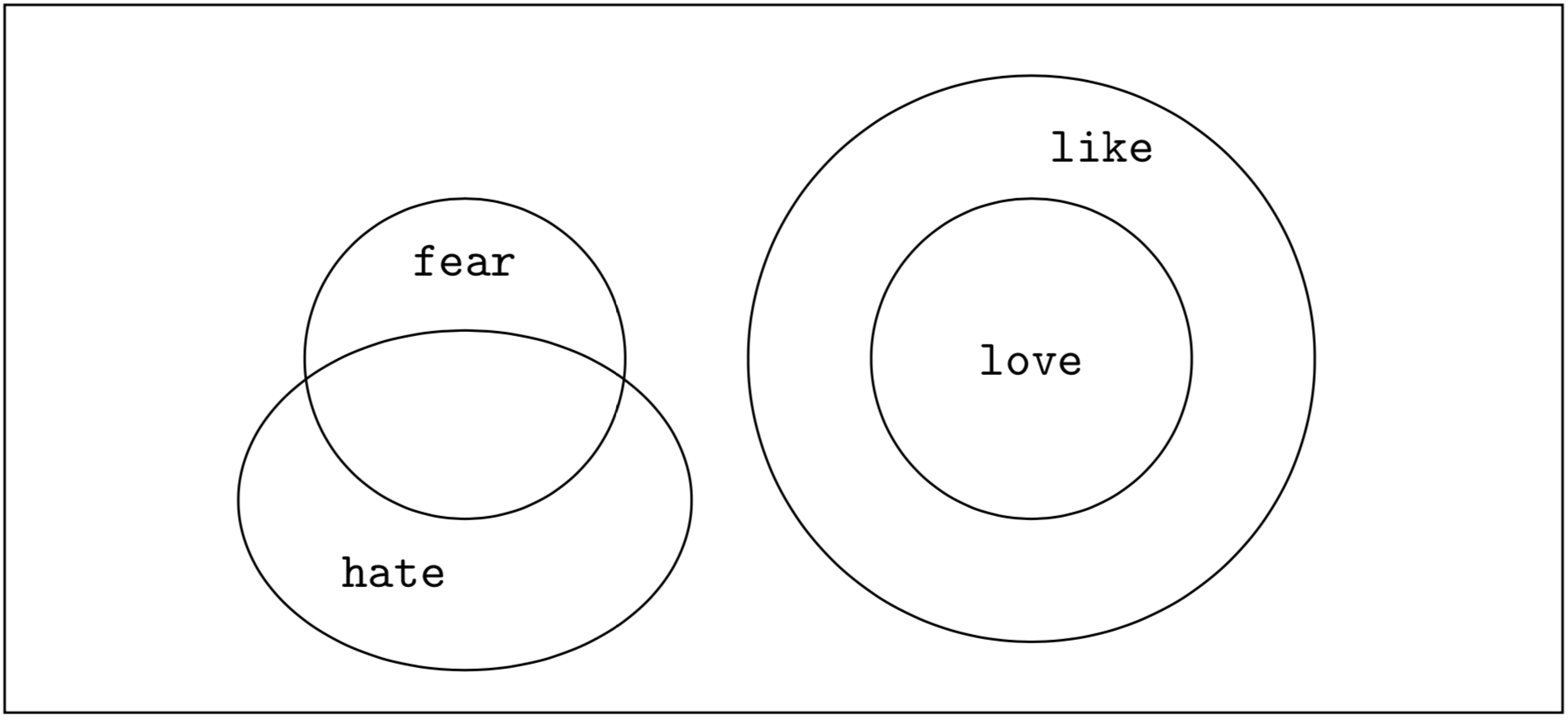}
\caption{}
\label{fig:verbs_new}
\end{subfigure}
\caption{Venn diagrams visualizing the taxonomy of (a) nouns $\mathcal{N}^{\mathcal{L}}$ and (b) verbs $\mathcal{V}^{\mathcal{L}}$ in $\mathcal{L}$.}
\label{fig:nouns-verbs}
\end{figure}

Let  $\mathcal{L}$ be the artificial language. Its vocabulary consists of four classes: quantifiers, nouns, (transitive) verbs and adverbs, represented by $\mathcal{Q}^{\mathcal{L}}$, $\mathcal{N}^{\mathcal{L}}$, $\mathcal{V}^{\mathcal{L}}$, $\mathcal{A}^{\mathcal{L}}$, respectively.
Lexical meanings of nouns and verbs are captured by a taxonomy of terms, as visualized in the Venn diagrams of Figure \ref{fig:nouns-verbs}. $\mathcal{Q}^{\mathcal{L}}$ contains the quantifiers \texttt{all} and \texttt{some}. $\mathcal{A}^{\mathcal{L}}$ includes the adverbs \texttt{not} and $\epsilon$ (the empty string).
Sentences in $\mathcal{L}$ can be generated according to the phrase structure grammar of Table \ref{tab:grammar_new}.

\begin{table}[ht]
\centering
\begin{tabular}{l|l|l|l|l}
S $\to$ NP VP   & Det $\to$ Adv Quant &
NP $\to$ Det NP & Quant $\to \mathcal{Q^{\mathcal{L}}}$ &
NP $\to$ Adv N \\ N $\to \mathcal{N^{\mathcal{L}}}$ &
VP $\to$ VP NP & V $\to \mathcal{V^{\mathcal{L}}}$ &
VP $\to$ Adv V & Adv $\to \mathcal{A^{\mathcal{L}}}$
\end{tabular}
\caption{Phrase structure grammar for artificial language $\mathcal{L}$.}
\label{tab:grammar_new}
\end{table}



\paragraph{Entailment relations}

Following \cite{maccartney2009extended} and \cite{bowman2014recursive}, seven different entailment relations are distinguished, as defined in Table \ref{tab:entailment-rels-table}. The relations are defined with respect to pairs of sets, but they are also applied to pairs of sentences, which can be interpreted as the sets of possible worlds where they hold true.

\begin{figure}[h]
\begin{minipage}{.45\linewidth}
\begin{table}[H]
\small
\centering
\caption{The seven entailment relations of \cite{maccartney2009extended}. $\mathcal{D}$ denotes the universe of discourse. \\
}
\label{tab:entailment-rels-table}
\setlength\tabcolsep{2pt}%
\begin{tabular}{cll}
\toprule 
& relation      & set-theoretic definition
\\ \midrule
\scriptsize{$\lor$} & cover   & $x \cap y \neq \emptyset \wedge x \cup y = \mathcal{D}$ \\
$^\wedge$ & negation               & $x \cap y = \emptyset \wedge x \cup y = \mathcal{D}$    \\
$< $ & forward entailment   & $x \subset y$                                           \\
$=$ & equivalence              & $x = y$                                                 \\
$ > $ & backward entailment  & $x \supset y$                                           \\
$\mid$  & alternation               & $x \cap y = \emptyset \wedge x \cup y \neq \mathcal{D}$ \\
$\# $  & independence                & (else)         \\
\bottomrule
\end{tabular}
\end{table}
\end{minipage}
\hspace{0.5cm}
\begin{minipage}{.45\linewidth}
\raggedright
\begin{table}[H]
\centering
\caption{FOL axiom representations of lexical entailment relations. For definition of relations, see Table \ref{tab:entailment-rels-table}. \\}
\label{tab:matrix-to-axioms}
\small
\setlength\tabcolsep{2pt}%
\begin{tabular}{cc}
\toprule
        relation & axiom representation in FOL \\
        \midrule
        $A < B$ & $\{\forall x (A(x) \to B(x)) \}$\\
        $A > B$ & $\{\forall x (B(x) \to A(x)) \}$\\
        $A \mid B$ & 
        $\{\forall x (\neg (A(x) \wedge B(x)))$,\\&
        $\neg \forall x (A(x) \lor B(x))
        \}$\\
        $A ^\wedge B$ & 
         $\{\forall x (\neg (A(x) \wedge B(x)))$,\\&
        $\forall x (A(x) \lor B(x))
        \}$\\
        $A{\scriptstyle\lor}B$ & $\{\forall x (\neg A(x) \to B(x)) \}$\\
        \bottomrule
    \end{tabular}
\end{table}
\end{minipage}
\end{figure}

\paragraph{Theorem prover}

We generate random pairs of sentences according to the grammar of $\mathcal{L}$ (e.g., `all Germans love all Romans' and `some Europeans like some Italians'). We then annotate these pairs with one of the 7 logical relations using the combination of an automated theorem prover for FOL with equality, Prover9, and a model builder, Mace4, proposed by \citet{prover9-mace4}. If Prover9 does not manage to find a proof in time, Mace4 takes over. 
To find the correct entailment relations between sentence pairs, we provide the theorem prover with the FOL translations of the actual sentences and the relevant lexical entailment relations in $\mathcal{L}$ (the axioms)\footnote{
The speed of Prover9 and Mace4 rapidly decreases as the number of axioms grows, so it is essential to keep the set of constraints considered per derivation as limited as possible. Hence, before computing whether $\varphi \vdash_{A^{\mathcal{L}}} \psi$, the collection of axioms is filtered in such a way as to retain the minimal set of formulas that could possibly be used in the proof or refutation of this particular entailment. This is done by dismissing all axioms containing predicates that do not occur in either $\varphi$ or $\psi$. E.g., if $\varphi = \forall x (A(x) \to C(x))$ and $\psi = \forall x (B(x) \to D(x))$, then all constraints in $A^{\mathcal{L}}$ containing terms not in $\{A,B,C,D\}$ are omitted. As the first term in a FOL representation of a $\mathcal{L}$ sentence (c.q. $A$ and $B$) is always a noun from $\mathcal{N}^{\mathcal{L}}$, while the second term (c.q. $C$ and $D$) is always a verb from $\mathcal{V}^{\mathcal{L}}$, only those axioms are used that relate the noun predicates or the verb predicates of both sentences to each other. No axioms combining terms from $\mathcal{N}^{\mathcal{L}}$ and $\mathcal{V}^{\mathcal{L}}$ exist, so this is generally the case. Additionally, not only identical but also equivalent axioms are eliminated. That is, if e.g. $\forall x (A(x) \lor B(x))$ is already included, $\forall x (B(x) \lor A(x))$ is redundant and cannot be added as well. 
}. The FOL representations of the axioms are derived according to the mapping in Table \ref{tab:matrix-to-axioms}. 


The grammar of $\mathcal{L}$ allows for the expression of approximately 40 million unique sentence pairs. In the default training set, we use slightly fewer than 30,000 of these pairs. The test set contains approximately 5,000 pairs. Sentences occur at most once in the data. 
A small sample is shown in Figure~\ref{table-example-relations}. Appendix \ref{sec:app-dist} shows the distribution in the train and test set of the seven entailment relations. 

\begin{figure*}[h]
\begin{minipage}{\textwidth}
\begin{tabular}{rll}
\verb+<+ & all Europeans like some Italians & not some Italians not like some Europeans\\ 
\verb+v+ & all Germans not hate all not Italians & not all not Italians love some not Italians \\
\verb+#+ & all children not hate all Romans & all not Italians not fear all Romans \\
\verb+|+ & some not Europeans like all not Italians & not some not Italians like all not Italians \\
\verb+^+ & not all not Germans not fear all Europeans & not some not Germans fear all Europeans 
\end{tabular}
\end{minipage}

    \caption{Some example pairs of sentences and their logical relations.}
    \label{table-example-relations}
\end{figure*}

\section{Learning models}

\begin{figure}[h]
    \centering
    \includegraphics[width=0.7\linewidth]{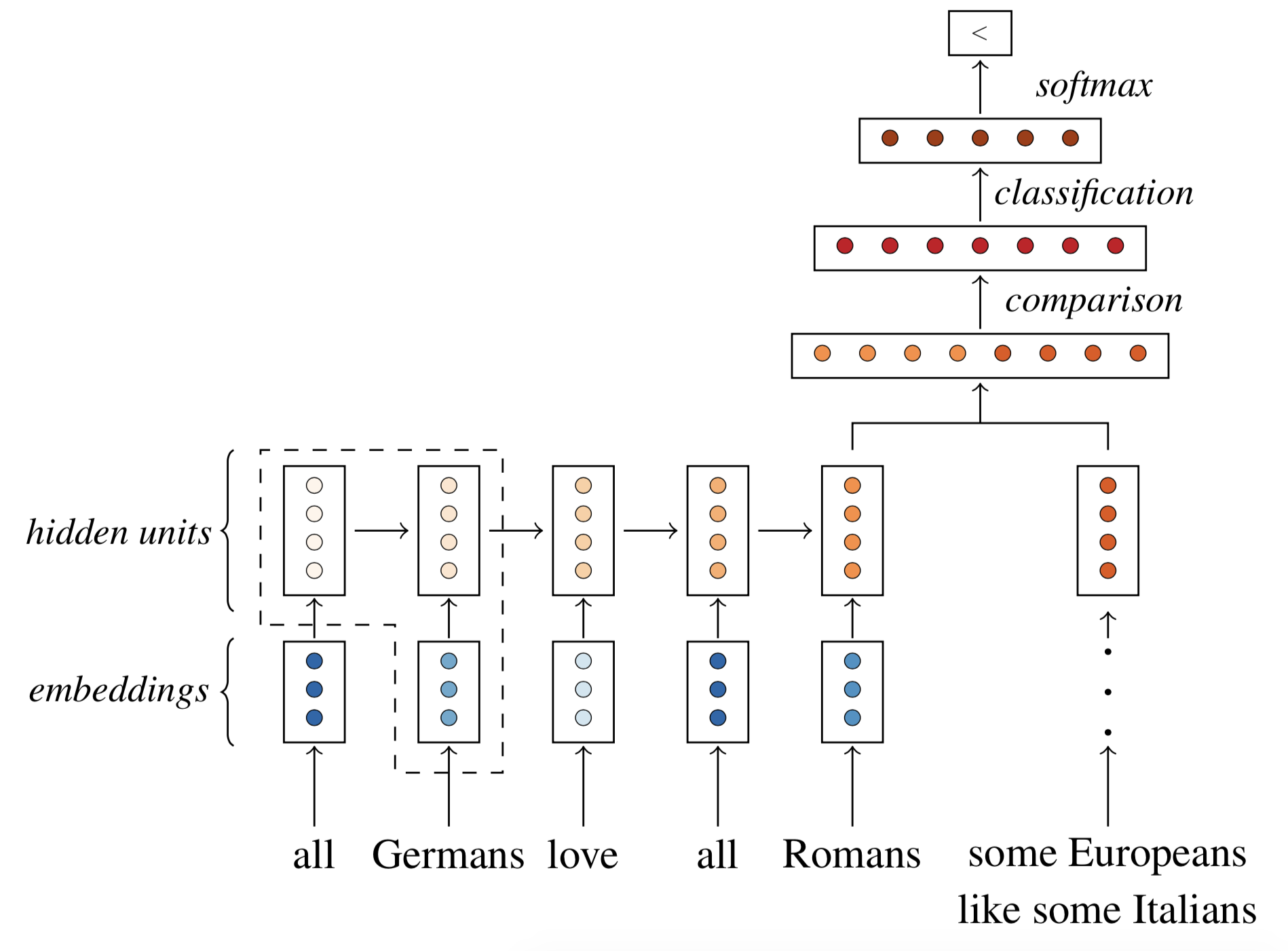}
    \caption{Visualization of the general recurrent model. The region in the dashed box represents any recurrent cell, which is repeatedly applied until the final sentence vector is returned.}
    \label{fig:rnn_diagram}
\end{figure}

Our main model is a recurrent network, sketched in Figure \ref{fig:rnn_diagram}. 
It is a so-called `Siamese' network because it uses the same parameters to process the left and the right sentence. The upper part of the model is identical to \citeauthor{bowman2014recursive}'s recursive networks. It consists of a comparison layer and a classification layer, after which a softmax function is applied to determine the most probable target class. The comparison layer takes the concatenation of two sentence vectors as input. The number of cells equals the number of words, so it differs per sentence. 

Our set-up resembles the Siamese architecture for learning sentence similarity of \cite{mueller2016siamese} and the LSTM classifier described in \cite{bowman2015large}. In the diagram, the dashed box indicates the location of an arbitrary recurrent unit. We consider SRN \citep{srn}, GRU \citep{gru} and LSTM \citep{lstm}.

\paragraph{Implementation details} 
Dimensionality of hidden units, word embeddings and comparison layers is 128, 25 and 75, respectively.
All recurrent networks have a single hidden layer. 
%
Prior to training, all hidden units are initialized as zero vectors. Network parameters are initialized by sampling from a uniform distribution; word embeddings by sampling from a normal distribution.
Weights of the recurrent units are drawn from the uniform distribution $\mathcal{U}(-1 / {\sqrt{h}}, 1 / {\sqrt{h}})$. In our case $h = 128$, so the lower bound is $-1 / {\sqrt{128}}$ and the upper bound $1/{\sqrt{128}}$. Non-recurrent parameters, belonging to the linear comparison and classification layers, are initialized uniformly randomly according to distribution $\mathcal{U}(-1 / \sqrt{fan_{in}},1 / \sqrt{fan_{in}})$, where $fan_{in}$ denotes the number of input units. The comparison layer is initialized according to $\mathcal{U}(-1/{\sqrt{50}}, 1/{\sqrt{50}})$, because its input is the concatenation of two 25-dimensional sentence vectors, and the initial classification layer weights are drawn from  $\mathcal{U}(-1/{\sqrt{75}}, 1/{\sqrt{75}})$, because the comparison layer outputs are 75-dimensional. AdaDelta \citep{adadelta} is used as optimizer. No dropout is applied. 


\paragraph{Baselines}
We consider three baselines used in earlier work by \citet{bowman2014recursive}: the recursive (tree-shaped) neural network (tRNN) and the recursive neural tensor network (tRNTN), which process the sentences according to their syntactic structure, and a simple bag-of-words model, implemented as a summing neural network based on a unweighted vector mixture model (sumNN). 

\section{Results}

Training and testing accuracies after 50 training epochs, averaged over five different model runs, are shown in Table \ref{tab:results}.
All recurrent models outperform the summing baseline. Even the simplest recurrent network, the SRN, achieves higher training and testing accuracy scores than the tree-shaped matrix model. The GRU and LSTM even beat the tensor model. The LSTM obtains slightly lower scores than the GRU, which is unexpected given its more complex design, but perhaps the current challenge does not require separate forget and input gates. 
For more insight into the types of errors made by the best-performing (GRU-based) model, we refer to the confusion matrices in Appendix \ref{sec:app-errors}.

\begin{figure}[h]
\begin{minipage}{.52\linewidth}
\begin{table}[H]
    \centering
           \caption{Training and testing accuracy scores on the FOL inference task. Mean and standard deviation over five runs. 
    }
    \label{tab:results}
    \begin{tabular}{lll}
    \toprule
                                 & \multicolumn{1}{c}{train} & \multicolumn{1}{c}{test} \\ \midrule
\multicolumn{1}{l}{sumNN} & $55.9 \pm 0.7$                       & $51.3 \pm 0.5$                     \\
\multicolumn{1}{l}{tRNN}  & $83.0 \pm 9.8$                       & $81.7 \pm 9.8$                     \\
\multicolumn{1}{l}{tRNTN} & $99.0 \pm 0.1$                      & $94.1 \pm 0.3$                 \\
\multicolumn{1}{l}{SRN}  & $97.9 \pm 0.2$  & $87.3 \pm 0.5$ \\
\multicolumn{1}{l}{GRU}  & $100.0 \pm 0.0$ & $96.0 \pm 0.6$\\
\multicolumn{1}{l}{LSTM} & $100.0 \pm 0.0$ & $94.6 \pm 2.0$\\
\bottomrule
\end{tabular}
 \end{table}
\end{minipage}
\hspace{0.5cm}
\begin{minipage}{.45\linewidth}
\raggedright

\begin{table}[H]
    \centering
    \caption{Accuracy scores on the FOL inference task for models trained on pairs of sentences with lengths 5, 7 or 8 and tested on pairs of sentences with lengths 6 or 9. Mean and standard deviation over five runs.}
    \label{tab:new-results-rnn-length-exp}
    \begin{tabular}{lll}
    \toprule
                                 & \multicolumn{1}{c}{train} & \multicolumn{1}{c}{test} \\ \midrule
\multicolumn{1}{l}{SRN} & $98.2 \pm 0.2$                       & $28.8 \pm 4.2$                    \\
\multicolumn{1}{l}{GRU}  & $100.0 \pm 0.0$                       & $90.4 \pm 1.2$                    \\
\multicolumn{1}{l}{LSTM} & $100.0 \pm 0.0$                       & $88.8 \pm 2.2$  \\                
\bottomrule
\end{tabular}
\end{table}

\end{minipage}
\end{figure}

The consistently higher testing accuracy provides evidence that the recurrent networks are not only capable of recognizing FOL entailment relations between unseen sentences. They can also outperform the tree-shaped models on this task, although they do not use any of the symbolic structure that seemed to explain the success of their recursive predecessors. The recurrent classifiers have learned to apply their own strategies, which we will investigate in the remainder of this paper.


\section{Zero-shot, compositional generalization}


Compositionality is the ability to interpret and generate a possibly infinite number of constructions from known constituents, and is commonly understood as one of the fundamental aspects of human learning and reasoning (\cite{chomsky1957syntactic, montague1970universal}). It has often been claimed that neural networks operate on a merely associative basis, lacking the compositional capacities to develop systematicity without an abundance of training data. See e.g. \cite{fodor1988connectionism}, \cite{marcus1998rethinking}, \cite{calvo2014architecture}. Especially recurrent models have recently been regarded quite sceptically in this respect, following the negative results established by \cite{lake2017building} and \cite{lake2017still}. Their research suggests that recurrent networks only perform well provided that there are no systematic discrepancies between train and test data, whereas human learning is robust with respect to such differences thanks to compositionality. 

In this section, we report more positive results on compositional reasoning of our Siamese networks. We focus on zero-shot generalization: correct classification of examples of a type that has not been observed before. 
Provided that atomic constituents and production rules are understood, compositionality does not require that abundantly many instances embodying a semantic category are observed. 
We will consider in turn what set-up is required to demonstrate zero-shot generalization to unseen lengths, and to generalization to sentences composed of novel words.

\subsection{Unseen lengths}

We test if our recurrent models are capable of generalization to unseen lengths. 
Neural models are often considered incapable of such generalization, allegedly because they are limited to the training space \citep{marcus2003algebraic, kaiser2015neural, reed2015neural, evans2018learning}. 
We want to test if this is the case for the recurrent models studied in this paper. The language $\mathcal{L}$ licenses a heavily constrained set of grammatical configurations, but it does allow the sentence length to vary according to the number of included negations. A perfectly compositional model should be able to interpret statements containing any number of negations, on condition that it has seen an instantiation at least once at each position where this is allowed. 

In a new experiment, we train the models on pairs of sentences with length 5, 7 or 8, and test on pairs of sentences with lengths 6 or 9. As before, the training and test sets contain some 30,000 and 5,000 sentence pairs, respectively. Results are shown in Table \ref{tab:new-results-rnn-length-exp}.

All recurrent models obtain (near-)perfect training accuracy scores. What happens on the test set is interesting. It turns out that the GRU and LSTM can generalize from lengths 5, 7 and 8 to 6 and 9 very well, while the SRN faces serious difficulties.
It seems that training on lengths 5, 7 and 8, and thereby skipping length 6, enables the GRU and LSTM to generalize to unseen sentence lengths 6 and 9. Training on lengths 5-7 and testing on lengths 8-9 yields low test scores for all models.
The GRU and LSTM gates appear to play a crucial role, because the results show that the SRN does not have this capacity at all.

\subsection{Unseen words}

In the next experiment, we assess whether our GRU-based model, which performed best in the preceding experiments, is capable of zero-shot generalization to sentences with novel words. 
The current set-up cannot deal with unknown words, so instead of randomly initializing an embedding matrix that is updated during training, we use pretrained, 50-dimensional GloVe embeddings \citep{pennington2014glove} that are kept constant. 
Using GloVe embeddings, the GRU model obtains a mean training accuracy of 100.0\% and a testing accuracy of 95.9\% (averaged over five runs). The best-performing model (with 100.0\% training and 97.1\% testing accuracy) is used in the following zero-shot experiments. 

\paragraph{Synonyms}

One of the most basic relations on the level of lexical semantics is synonymy, which holds between words with equivalent meanings. 
In the language $\mathcal{L}$, a word can be substituted with one of its synonyms without altering the entailment relation assigned to the sentence pairs that contain it. 
If the GRU manages to perform well on such a modified data set after receiving the pretrained GloVe embedding of the unseen word, this is a first piece of evidence for its zero-shot generalization skills. We test this for several pairs of synonymous words. The best-performing GRU is first evaluated with respect to the fragment of the test data containing the original word $w$, and consequently with respect to that same fragment after replacing the original word with its synonym $s(w)$. The pairs of words, the cosine distance $cos\_dist(w,s(w))$ between their GloVe embeddings and the obtained results are listed in Table \ref{tab:one-shot-synonyms}. 

\begin{table*}[ht]
    \centering
    \caption{Effect on best-performing GRU of replacing words $w$ by unseen synonyms $s(w)$ in the test set and providing the model with the corresponding GloVe embedding.}
    \label{tab:one-shot-synonyms}
    \begin{tabular}{llccc}
        \toprule
        \multicolumn{1}{c}{} & \multicolumn{1}{c}{} & & \multicolumn{2}{c}{test accuracy} \\ 
        \cmidrule{4-5}
        \multicolumn{1}{c}{$w$} & \multicolumn{1}{c}{$s(w)$} & $cos\_dist(w, s(w))$ & \multicolumn{1}{c}{containing $w$} & containing $s(w)$ \\ 
        \midrule
        \multicolumn{1}{l}{children} & \multicolumn{1}{l}{kids} & 0.21 & \multicolumn{1}{c}{98.3} & 91.9 \\
        \multicolumn{1}{l}{love} & \multicolumn{1}{l}{adore}  & 0.57 & \multicolumn{1}{c}{97.0} & 92.5 \\
        \multicolumn{1}{l}{fear} & \multicolumn{1}{l}{dread} & 0.39 & \multicolumn{1}{c}{97.2} & 91.3 \\
        \multicolumn{1}{l}{hate} & \multicolumn{1}{l}{detest} & 0.56 & \multicolumn{1}{c}{97.4} & 48.2 \\
        \bottomrule
\end{tabular}
\end{table*}

For the first three examples in Table \ref{tab:one-shot-synonyms}, substitution only decreases testing accuracy by a few percentage points. Apparently, the word embeddings of the synonyms encode the lexical properties that the GRU needs to recognize that the same entailment relations apply to the sentence pairs. This does not prove that the model has distilled essential information about hyponymy from the GloVe embeddings. It could also be that the word embeddings of the replacement words are geometrically very similar to the originals, so that it is an algebraic necessity that the same results arise. However, this suspicion is inconsistent with the result of changing `hate' into `detest'. The cosine distance between these words is 0.56, so according to this measure their vectors are more similar than those representing `love' and `adore' (which have a cosine distance of 0.57). Nonetheless, replacing `hate' with `detest' confuses the model, whereas substitution of `love' into `adore' only decreases testing accuracy by 4.5 percentage points. This illustrates that robustness of the GRU in this respect is not a matter of simple vector similarity. In those cases where substitution into synonyms does not confuse the model it must have recognized a non-trivial property of the new word embedding that licenses particular inferences. 

\paragraph{Ontological twins}
In our next experiment, we replace a word not by its synonym, but by a word that has the same semantics in the context of artificial language $\mathcal{L}$. 
We thus consider pairs of words that can be substituted with each other without affecting the entailment relation between any pair of sentences in which they feature.  We call such terms `ontological twins'. 
Technically, if $\odot$ is an arbitrary lexical entailment relation and $\mathcal{O}$ is an ontology, then $w$ and $v$ are ontological twins if and only if $w, v \in \mathcal{O}$ and for all $u \in \mathcal{O}$, if $u \not \in \{ w,v \} $ then $w \odot u \Leftrightarrow v \odot u$. This trivially applies to self-identical terms or synonyms, but in the strictly defined hierarchy of $\mathcal{L}$ it is also the case for pairs of terms $w, v$ that maintain the same lexical entailment relations to all other terms in the taxonomy.

Examples of ontological twins in the taxonomy of nouns $\mathcal{N}^{\mathcal{L}}$ are `Romans' and `Venetians' . This can easily be verified in the Venn diagram of Figure \ref{fig:nouns_new} by replacing `Romans' with `Venetians' and observing that the same hierarchy applies. The same holds for e.g. `Germans' and `Polish' or for `children' and `students'. For several such word-twin pairs the GRU is evaluated with respect to the fragment of the test data containing the original word $w$, and with respect to that same fragment after replacing the original word with ontological twin $t(w)$. Results are shown in Table \ref{tab:one-shot-twins}. 

\begin{table*}[h]
    \centering
    \caption{Effect on best-performing GRU of replacing words $w$ by unseen ontological twins $t(w)$ in the test set and providing the model with the corresponding GloVe embedding.}
    \label{tab:one-shot-twins}
    \begin{tabular}{llccc}
        \toprule
        \multicolumn{1}{c}{}          & \multicolumn{1}{c}{}             &                          & \multicolumn{2}{c}{test accuracy} \\ 
        \cmidrule{4-5}
        \multicolumn{1}{c}{$w$}      & \multicolumn{1}{c}{$t(w)$}      & $cos\_dist(w, t(w))$ & \multicolumn{1}{c}{containing $w$}             & containing $t(w)$             \\ 
        \midrule
        \multicolumn{1}{l}{}         & \multicolumn{1}{l}{Venetians}   & 0.28                     & \multicolumn{1}{c}{}                           & 97.3                          \\
        \multicolumn{1}{l}{Romans}   & \multicolumn{1}{l}{Milanese}    & 0.72                     & \multicolumn{1}{c}{97.3}                       & 95.4                          \\
        \multicolumn{1}{l}{}         & \multicolumn{1}{l}{Neapolitans} & 0.57                     & \multicolumn{1}{c}{}                           & 95.4                          \\ \hline
        \multicolumn{1}{l}{}         & \multicolumn{1}{l}{Polish}      & 0.37                     & \multicolumn{1}{c}{}                           & 96.3                          \\
        \multicolumn{1}{l}{Germans}  & \multicolumn{1}{l}{Dutch}       & 0.40                     & \multicolumn{1}{c}{96.8}                       & 78.3                          \\
        \multicolumn{1}{l}{}         & \multicolumn{1}{l}{Spanish}     & 0.50                     & \multicolumn{1}{c}{}                           & 62.9                          \\ \hline
        \multicolumn{1}{l}{} & \multicolumn{1}{l}{students}    & 0.27                     & \multicolumn{1}{c}{}                       & 94.6                          \\
        \multicolumn{1}{l}{children}         & \multicolumn{1}{l}{women}       & 0.24                     & \multicolumn{1}{c}{98.3}                           & 91.8                          \\
        \multicolumn{1}{l}{}         & \multicolumn{1}{l}{linguists}   & 0.92                     & \multicolumn{1}{c}{}                           & 86.3\\
        \bottomrule
\end{tabular}
\end{table*}

The examples in Table \ref{tab:one-shot-twins} suggest that the best-performing GRU is largely robust with respect to substitution into ontological twins. Replacing `Romans' with other urban Italian demonyms hardly affects model accuracy on the modified fragment of the test data. As before, there appears to be no correlation with vector similarity because the cosine distance between the different twin pairs has a much higher variation than the corresponding accuracy scores. `Germans' can be changed into `Polish' without significant deterioration, but substitution with `Dutch' greatly decreases testing accuracy. The situation is even worse for `Spanish'. Again, cosine similarity provides no explanation - `Spanish' is still closer to `Germans' than `Neapolitans' to `Romans'. Rather, the accuracy appears to be negatively correlated with the geographical distance between the national demonyms. After replacing `children' with `students', `women' or `linguists', testing scores are still decent. 

\paragraph{Alternative hierarchies}

So far, we replaced individual words in order to assess whether the GRU can generalize from the vocabulary to new notions that have comparable semantics in the context of this entailment recognition task. The examples have illustrated that the model tends to do this quite well. In the last zero-shot learning experiment, 
we replace sets of nouns instead of single words, in order to assess the flexibility of the relational semantics that our networks have learned.
Formally, the replacement can be regarded as a function $r$, mapping words $w$ to substitutes $r(w)$. Not all items have to be replaced. For an ontology $\mathcal{O}$, the function $r$ must be such that for any $w, v \in \mathcal{O}$ and lexical entailment relation $\odot$, $w \odot v \Leftrightarrow r(w) \odot r(v)$. The result of applying $r$ can be called an `alternative hierarchy'. 

An example of an alternative hierarchy is the result of the replacement function $r_1$ that maps `Romans' to `Parisians' and `Italians' to `French'. Performing this substitution in the Venn diagram of Figure \ref{fig:nouns_new} shows that the taxonomy remains structurally intact. The best-performing GRU is evaluated on the fragment of the test data containing `Romans' or `Italians', and consequently on the same fragment after implementing replacement $r_1$ and providing the model with the GloVe embeddings of the unseen words. Replacement $r_1$ is incrementally modified up until replacement $r_4$, which substitutes all nouns in $\mathcal{N}^{\mathcal{L}}$. The results of applying $r_1$ to $r_4$ are shown in Table \ref{tab:one-shot-hierarchies}. 

\begin{table*}[h]
    \centering
    \caption{Effect on best-performing GRU of replacing noun ontology $\mathcal{N}^{\mathcal{L}}$ with alternative hierarchies as per the replacement functions $r_1$ to $r_4$. Vertical dots indicate that cell entries do not change on the next row.}
    \label{tab:one-shot-hierarchies}
    \begin{tabular}{lcccccccccccc}
        \toprule
        &               &               &                &               &                &              &               &               &                &               & \multicolumn{2}{c}{\begin{tabular}[c]{@{}c@{}}test accuracy \end{tabular}} \\ 
        \cmidrule{12-13} 
        \multicolumn{1}{l}{}      & \multicolumn{2}{c}{Romans} & \multicolumn{2}{c}{Italians}    & \multicolumn{2}{c}{Germans}  & \multicolumn{2}{c}{Europeans} & \multicolumn{2}{c}{children} & \multicolumn{1}{c}{\begin{tabular}[c]{@{}c@{}}before \\ substitution\end{tabular}} & \begin{tabular}[c]{@{}c@{}}after \\ substitution\end{tabular} \\ 
        \midrule
        \multicolumn{1}{l}{$r_1$} & \multicolumn{2}{c}{Parisians}   & \multicolumn{2}{c}{French} & \multicolumn{2}{c}{$\vdots$} & \multicolumn{2}{c}{$\vdots$} & \multicolumn{2}{c}{$\vdots$}  & \multicolumn{1}{c}{97.2}                         & 93.5                        \\
        \multicolumn{1}{l}{$r_2$} & \multicolumn{2}{c}{$\vdots$} & \multicolumn{2}{c}{$\vdots$}  & \multicolumn{2}{c}{Polish}   & \multicolumn{2}{c}{$\vdots$} & \multicolumn{2}{c}{$\vdots$}  & \multicolumn{1}{c}{97.1}                         & 93.9                        \\
        \multicolumn{1}{l}{$r_3$} & \multicolumn{2}{c}{$\vdots$} & \multicolumn{2}{c}{$\vdots$}  & \multicolumn{2}{c}{$\vdots$} & \multicolumn{2}{c}{Eurasians} & \multicolumn{2}{c}{$\vdots$}  & \multicolumn{1}{c}{97.1}                         & 87.6                        \\
        \multicolumn{1}{l}{$r_4$} & \multicolumn{2}{c}{$\vdots$} & \multicolumn{2}{c}{$\vdots$}  & \multicolumn{2}{c}{$\vdots$} & \multicolumn{2}{c}{$\vdots$} & \multicolumn{2}{c}{students} & \multicolumn{1}{c}{97.1}                         & 86.7\\
        \bottomrule
    \end{tabular}
\end{table*}

The results are positive: the GRU obtains 86.7\% accuracy even after applying $r_4$, which substitutes the entire ontology $\mathcal{N}^{\mathcal{L}}$ so that no previously encountered nouns are present in the test set anymore, although the sentences remain thematically somewhat similar to the original sentences. Testing scores are above 87\% for the intermediate substitutions $r_1$ to $r_3$. This outcome clearly shows that the classifier does not depend on a strongly customized word vector distribution in order to recognize higher-level entailment relations. Even if all nouns are replaced by alternatives with embeddings that have not been witnessed or optimized beforehand, the model obtains a high testing accuracy. This establishes obvious compositional capacities, because familiarity with structure and information about lexical semantics in the form of word embeddings are enough for the model to accommodate configurations of unseen words. 

What happens when we consider ontologies that have the same structure, but are thematically very different from the original ontology? Three such alternative hierarchies are considered: $r_{animals}$, $r_{religion}$ and $r_{America}$. Each of these functions relocalizes the noun ontology in a totally different domain of discourse, as indicated by their names. Table \ref{tab:one-shot-hierarchies-alternative} specifies the functions and their effect. 

\begin{table*}[h]
    \centering
    \caption{Effect on best-performing GRU of replacing noun ontology $\mathcal{N}^{\mathcal{L}}$ with alternative hierarchies as per the replacement functions $r_{animals}$, $r_{religion}$ and $r_{America}$. Accuracy is measured on the test set after applying the respective replacement functions.}
    \label{tab:one-shot-hierarchies-alternative}
    \begin{tabular}{lcccccc}
        \toprule
        & Romans       & Italians    & Germans      & Europeans  & children & \textit{accuracy} \\ 
        \midrule
        $r_{animals}$  & rabbits      & rodents     & cats         & mammals    & pets     & 59.0     \\
        $r_{religion}$ & calvinists   & protestants & catholics    & christians & orthodox & 56.3     \\
        $r_{America}$  & Clevelanders & Ohioans     & Californians & Americans  & women    & 58.2 \\
        \bottomrule
\end{tabular}
\end{table*}

Testing accuracy decreases drastically, which indicates that the model is sensitive to the changing topic. Variation between the scores obtained after the three transformations is limited. Although they are much lower than before, they are still far above chance level for a seven-class problem. This suggests that the model is not at a complete loss as to the alternative noun hierarchies. Possibly, including a few relevant instances during training could already improve the results.

\section{Discussion \& Conclusions}

We established that our Siamese recurrent networks (with SRN, GRU or LSTM cells) are able to recognize logical entailment relations without any a priori cues about syntax or semantics of the input expressions. Indeed, some of the recurrent set-ups even outperform tree-shaped networks, whose topology is specifically designed to deal with such tasks. 
This indicates that recurrent networks can develop representations that can adequately process a formal language with a nontrivial hierarchical structure. 
The formal language we defined did not exploit the full expressive power of first-order predicate logic; nevertheless by using standard first-order predicate logic, a standard theorem prover, and a set-up where the training set only covers a tiny fraction of the space of possible logical expressions, our experiments avoid the problems observed in earlier attempts to demonstrate logical reasoning in recurrent networks.

The experiments performed in the last few sections moreover show that the GRU and LSTM architectures exhibit at least basic forms of compositional generalization. In particular, the results of the zero-shot generalization experiments with novel lengths and novel words cannot be explained with a `memorize-and-interpolate' account, i.e. an account of the working of deep neural networks that assumes all they do is store enormous training sets and generalize only locally.  These results are relevant pieces of evidence in the decades-long debate on whether or not connectionist networks are fundamentally able to learn compositional solutions. Although we do not have the illusion that our work will put this debate to an end, we hope that it will help bring deep learning enthusiasts and skeptics a small step closer.



\bibliography{molthesis}
\bibliographystyle{acl_natbib}

\newpage
\appendix

\section{Appendices}

\subsection{Class distribution}
\label{sec:app-dist}

\begin{figure}[h]
    \centering
    \includegraphics[width=0.75\linewidth]{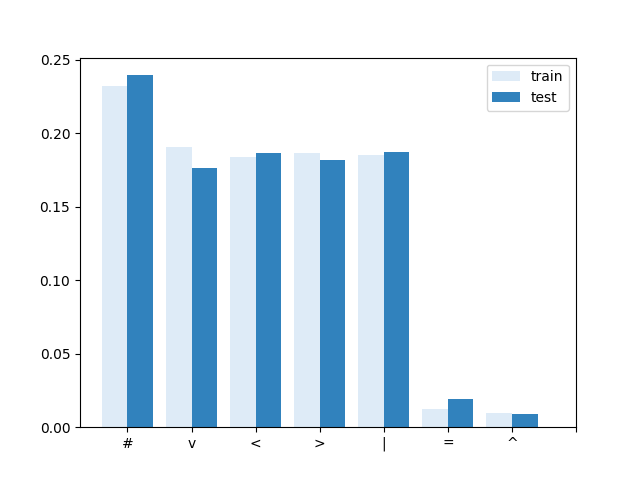}
    \caption{Histogram showing the relative frequency of each entailment relation in the train and test set.}
    \label{fig:data-dist}
\end{figure}

\subsection{Error statistics}
\label{sec:app-errors}

\begin{figure}[h]
\centering
\begin{subfigure}{0.49\textwidth}
\includegraphics[width=\textwidth]{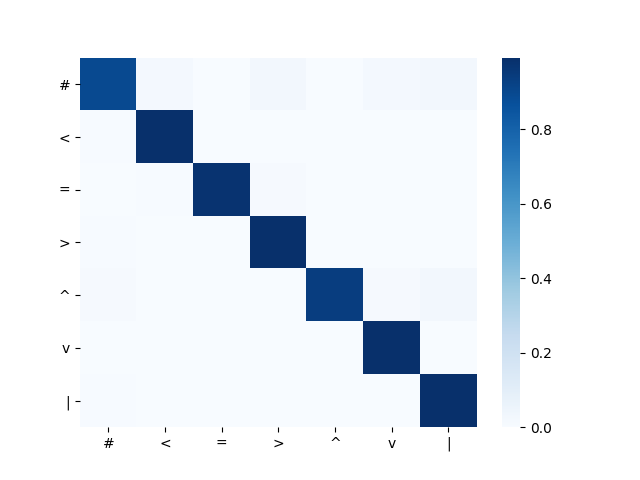}
\caption{}
\label{fig:conf-gru}
\end{subfigure}
\begin{subfigure}{0.49\textwidth}
\includegraphics[width=\textwidth]{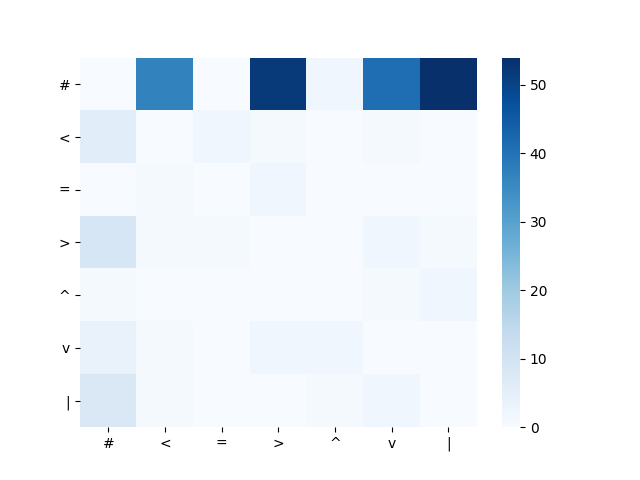}
\caption{}
\label{fig:conf-gru-onlyerrors}
\end{subfigure}
\caption{Confusion matrices of the best-performing GRU with respect to the test set. Rows represent targets, columns predictions. (a) row-normalized results for all test instances. (b) unnormalized results for misclassified test instances. Clearly, most errors are due to unrecognized or wrongly attributed independence.}
\label{fig:conf}
\end{figure}

\end{document}